%% file: main.tex
\documentclass[journal]{IEEEtran}

\usepackage[utf8]{inputenc} 
\usepackage{cite}
\usepackage{graphicx}
\usepackage{dblfloatfix}    
\usepackage{amsmath}
\usepackage{amssymb}
\usepackage{float}
\usepackage{booktabs}
\usepackage{array}
\usepackage{multirow}
\usepackage{xspace}
\usepackage{xcolor}         
\usepackage{url}            
\usepackage{hyperref}       

\definecolor{rseditlightblue}{rgb}{0.3,0.65,1}

\newcommand{\rsedit}{\textcolor{rseditlightblue}{RSEdit}\xspace}

\newcommand{\rseditdit}{\textcolor{rseditlightblue}{RSEdit-DiT}\xspace}

\newcommand{\rseditsubunet}{\textcolor{rseditlightblue}{RSEdit\textsubscript{U-Net}}\xspace}
\newcommand{\rseditsubdit}{\textcolor{rseditlightblue}{RSEdit\textsubscript{DiT}}\xspace}
\newcommand{\rseditsubditplus}{\textcolor{rseditlightblue}{RSEdit\textsubscript{DiT+}}\xspace}

\begin{document}

\title{RSEdit: Text-Guided Image Editing for Remote Sensing}

\author{Zhenyuan~Chen,
        Zechuan~Zhang,
        and~Feng~Zhang%
\thanks{Z. Chen and F. Zhang are with the School of Earth Sciences, Zhejiang University, Hangzhou, 310058 China, and Zhejiang Provincial Key Laboratory of Geographic Information Science. F. Zhang is also with Key Laboratory of Spatio-temporal Information and Intelligent Services (LSIIS), Ministry of Natural Resources of the People's Republic of China. Z. Zhang is with ReLER, CCAI, Zhejiang University, Hangzhou, China.}%
}


\maketitle

\begin{abstract}
In this paper, we explore text-guided image editing in the remote sensing domain using generative modeling. We propose \rsedit, a collection of models from U-Net to DiT with various configurations. Specifically, we present the first comprehensive study of conditioning strategies for building image editing models from off-the-shelf text-to-image ones. Our experiments show that \rsedit achieves the best instruction-faithful edits while preserving geospatial structure. We release the code at \url{https://github.com/Bili-Sakura/RSEdit-Preview} and checkpoints at \url{https://huggingface.co/collections/BiliSakura/rsedit}.
\end{abstract}


\begin{figure}[!t]
  \centering
  \includegraphics[width=\columnwidth]{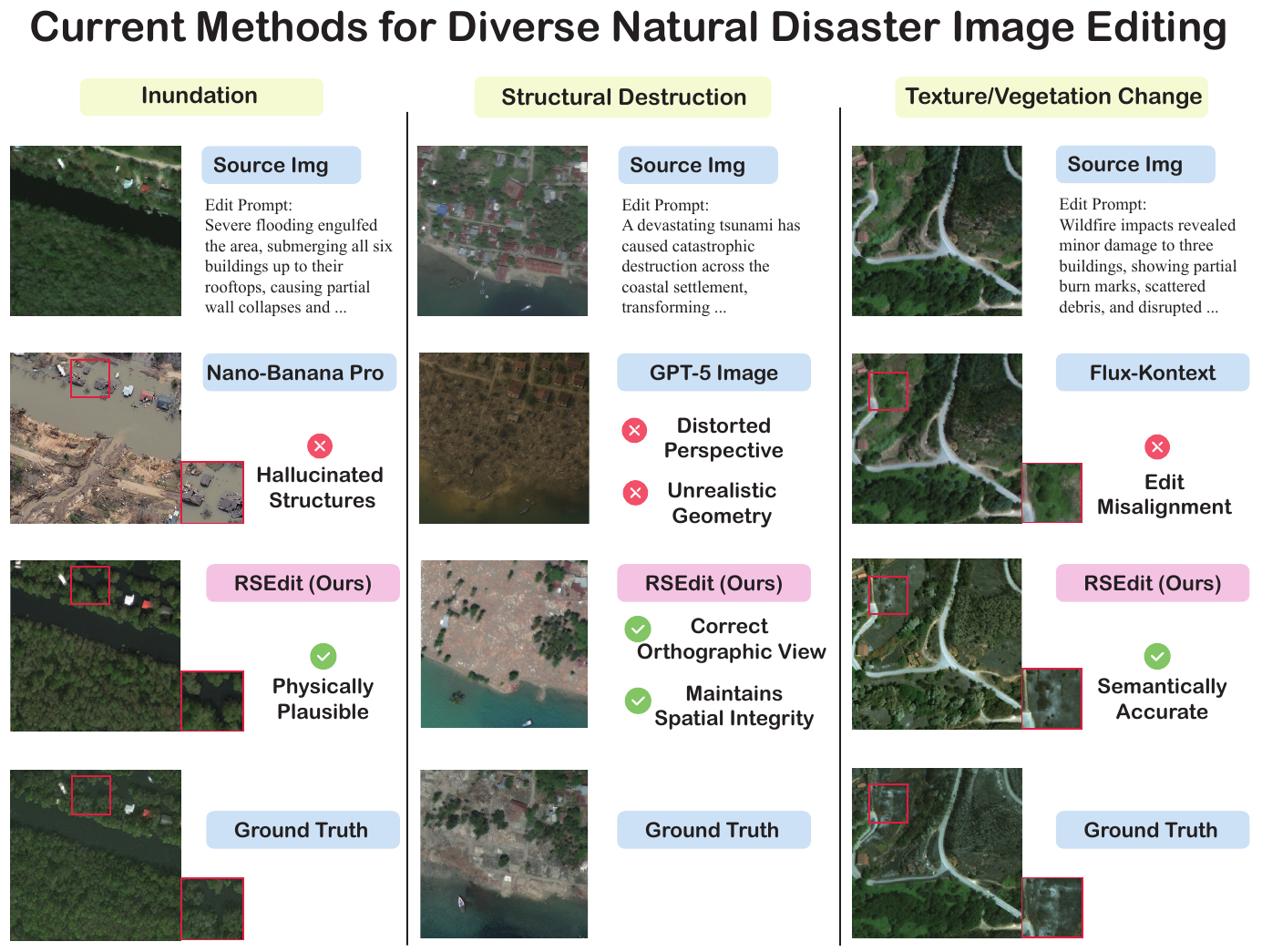}
  \caption{Illustration of text-guided remote sensing image editing with \rsedit{}.}
  \label{fig:teaser_rsedit}
\end{figure}

\input{sections/introduction}

\input{sections/methods}
\input{sections/results}

\input{sections/conclusion}

\bibliographystyle{IEEEtran}
\bibliography{references}

\end{document}

%% file: sections/introduction.tex
\section{Introduction}\label{sec:intro}

Text-guided image editing has progressed rapidly with diffusion models, enabling instruction-following edits in natural images~\cite{brooks2023instructpix2pix}. In remote sensing (RS), such editing could serve as a practical data engine: modeling disaster events with scarce imagery limits diversity for training robust analysis models, whereas a specialized editor can synthesize disaster scenarios on demand from any pre-event image---e.g., transforming a village into a post-flood landscape with controllable damage severity. The utility spans disaster simulation, urban-change analysis, and seasonal scenario generation. However, directly applying general editors to satellite imagery is unreliable: they often hallucinate structures, break orthographic geometry, or over-edit unchanged regions.

RS generation has advanced through text-conditional and image-conditional models~\cite{liuText2EarthUnlockingTextdriven2025,zhengChangen2MultiTemporalRemote2024}; however, these methods mainly focus on synthesis rather than instruction-based editing of existing imagery. General-domain editors~\cite{mengSDEditGuidedImage2022} achieve strong visual quality, yet lack RS world knowledge. Early RS editing~\cite{hanExploringTextGuidedSingle2025a,zhaoChangeBridgeSpatiotemporalImage2025} relies on constrained architectures or auxiliary segmentation supervision, limiting capability on long, semantically rich prompts. To fill this gap, we propose \rsedit, a collection of image editing models specialized for remote sensing images.

\noindent\textbf{Contributions.} (1) A comprehensive architecture--conditioning study for RS image editing, covering U-Net/DiT backbones and channel/token concatenation strategies under a unified experimental setting. (2) We further explore the performance gap between architectural designs and build a tailored \rseditsubditplus{} variant that matches the performance of the U-Net variant with window attention.

%% file: sections/methods.tex
\section{Methods}\label{sec:methods}

\subsection{Preliminaries}
We formulate text-guided remote sensing image editing as a conditional generation problem. Given a source satellite image $I \in \mathbb{R}^{H \times W \times 3}$ and a text instruction $T$ describing the desired semantic change (e.g., ``flood the coastal area''), the goal is to generate an edited image $I'$ that follows the instruction while preserving the spatial structure and unedited content of $I$.

We operate within the latent diffusion framework~\cite{rombachHighResolutionImageSynthesis2022,hoDenoisingDiffusionProbabilistic2020}. The model learns to denoise a latent code $z_t \in \mathbb{R}^{h \times w \times d_z}$ at timestep $t$ conditioned on a text embedding $c_T$ and an image condition $c_I$. The training objective is:
\begin{equation}
    \mathcal{L} = \mathbb{E}_{z_0,t,\epsilon,I,T}\left[\left\|\epsilon-\epsilon_\theta(z_t,t,c_T,c_I)\right\|_2^2\right].
\end{equation}
The core challenge is designing $c_I$ so that the conditioning signal aligns with the inductive bias of the underlying backbone---convolutional (U-Net) versus sequence-based (DiT). General-domain editors often use conditioning schemes optimized for one architecture; adapting to RS data requires a flexible strategy that respects both convolutional and transformer priors.

\subsection{Architecture-Aware Conditioning}
\begin{figure}[t]
  \centering
  \newsavebox{\conditioningablationcropbox}%
  \sbox{\conditioningablationcropbox}{\includegraphics[page=2]{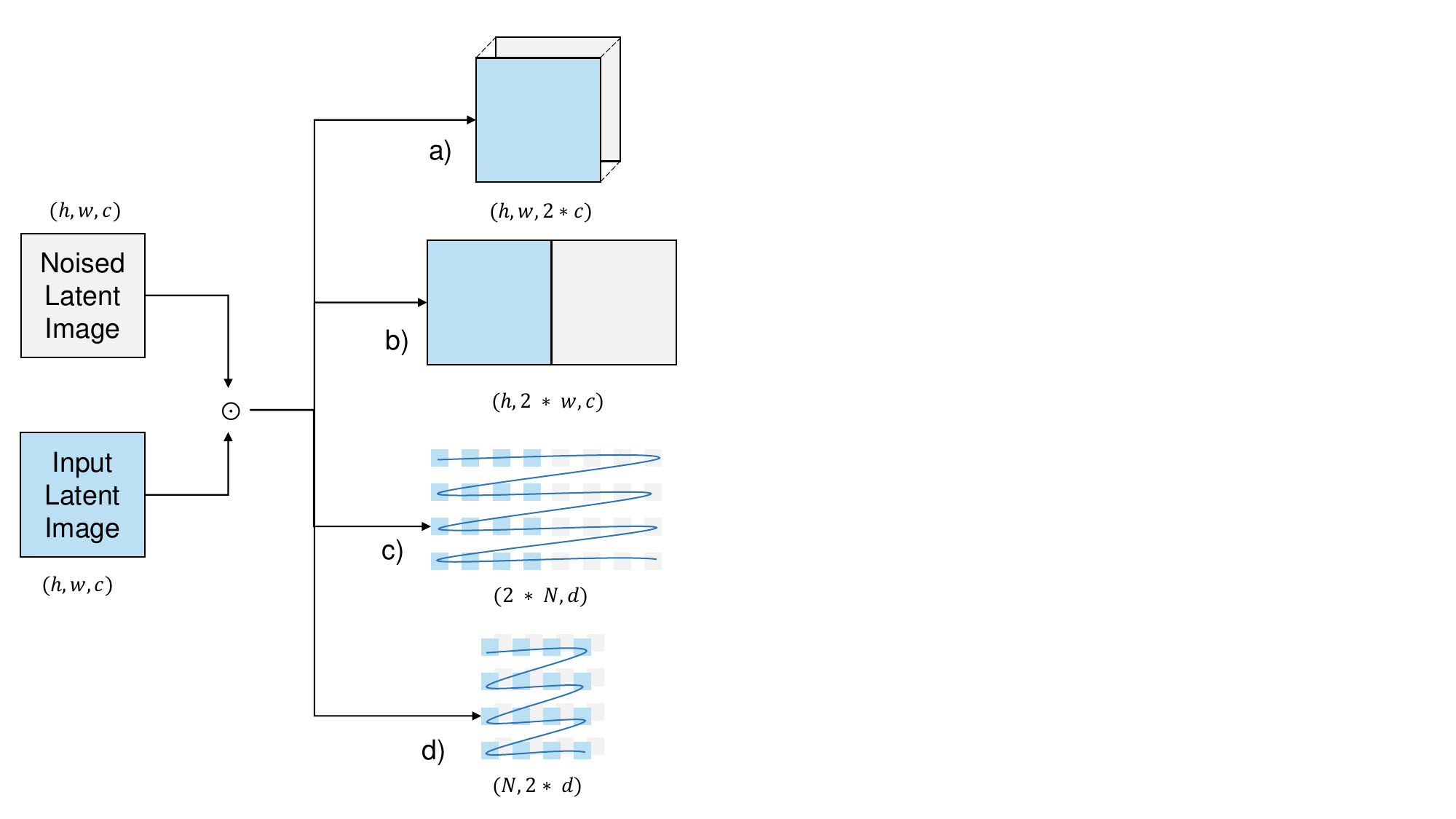}}%
  \includegraphics[width=\linewidth,trim=0 0 {0.5\wd\conditioningablationcropbox} 0,clip,page=2]{figures/figuresv2.pdf}
  \caption{Architecture-aware conditioning in \rsedit. U-Net: (a)~channel concatenation, (b)~token concatenation. DiT: (c)~conditioning with token layout $(2N,d)$; (d)~conditioning with token layout $(N,2d)$. Curves denote the DiT token flatten order, which is handled by 2D position embedding.}
  \label{fig:conditioning_ablation}
\end{figure}

Existing editing frameworks are often tied to architecture-specific heuristics, limiting transfer to diverse backbones. We argue that a robust RS editor must be \textit{architecture-aware}: the efficacy of the conditioning signal depends intrinsically on the inductive bias of the underlying diffusion backbone. \rsedit pairs each backbone with concrete layouts for feeding the VAE-encoded source and the noisy latent $z_t$ into the denoiser. Fig.~\ref{fig:conditioning_ablation} sketches these layouts without a separate equation block.

For convolutional U-Nets~\cite{ronnebergerUnetConvolutionalNetworks2015, rombachHighResolutionImageSynthesis2022}, we build on InstructPix2Pix-style spatial alignment~\cite{brooks2023instructpix2pix}. Figure~\ref{fig:conditioning_ablation} shows the U-Net side (panels~(a) and~(b)) of that design space: panel~(a) widens feature maps by channel-concatenating the source latent with $z_t$ before convolutional blocks, whereas panel~(b) concatenates separately patchified source and noisy streams at the token level. Panels~(c) and~(d) depict DiT conditioning with token layouts $(2N,d)$ and $(N,2d)$, respectively. For Diffusion Transformers (DiTs)~\cite{peeblesScalableDiffusionModels2023,chenPixArtaFastTraining2024}, we consider the analogous token-level versus channel-level fusion under the same latent-diffusion editing setup.

\subsection{\rseditsubditplus{} Architecture}
\begin{figure}[t]
  \centering
  \newsavebox{\rfreceptivefieldcropbox}%
  \sbox{\rfreceptivefieldcropbox}{\includegraphics[page=5]{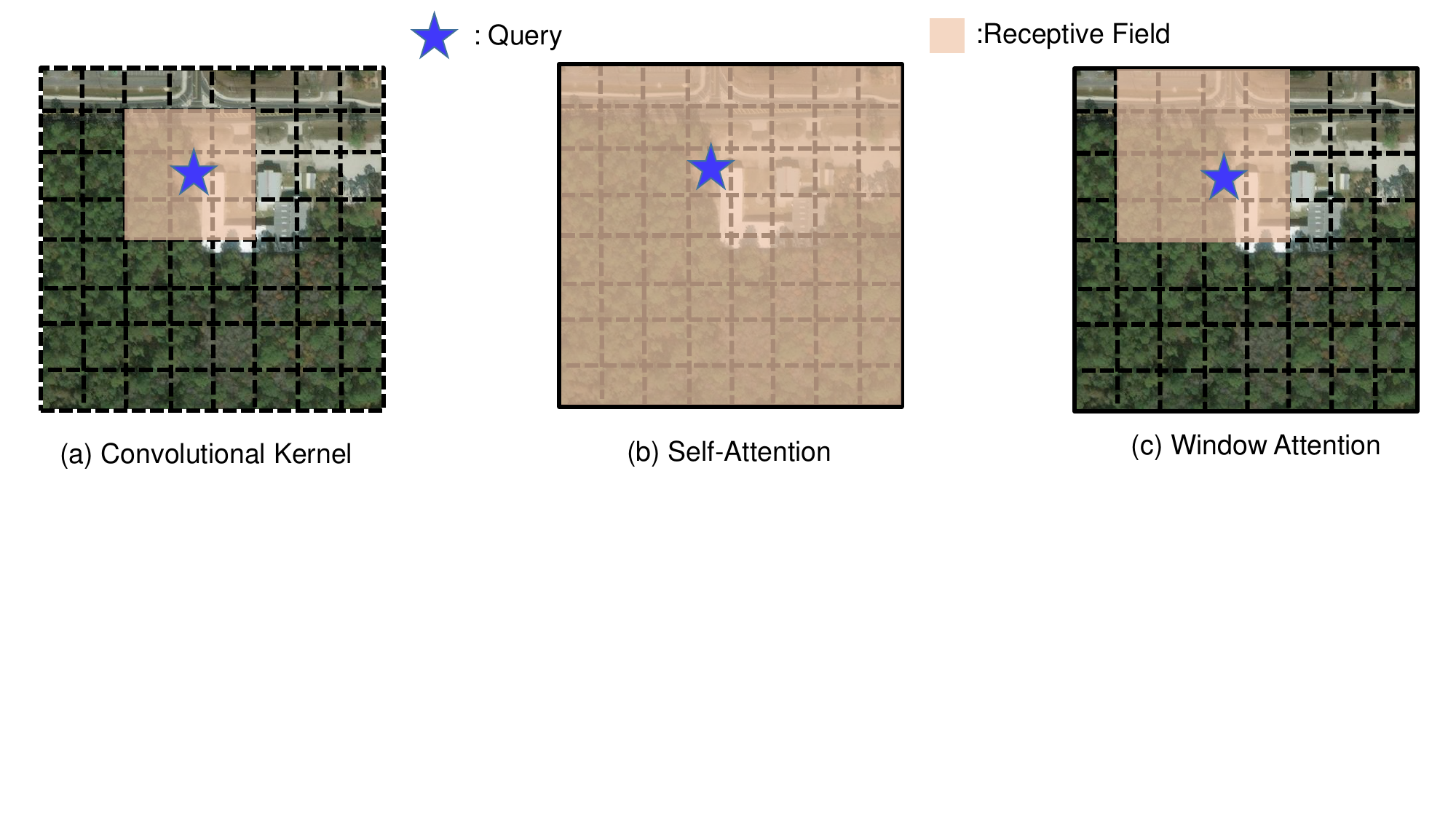}}%
  \includegraphics[width=\linewidth,trim=0 {0.4\ht\rfreceptivefieldcropbox} 0 0,clip,page=5]{figures/figuresv2.pdf}
  \caption{Receptive field comparison: (a)~convolutional kernel, (b)~full self-attention, (c)~window attention. Shaded regions indicate context seen from the query (star). Window attention balances the compute budget and receptive-field extent between these extremes.}
  \label{fig:receptive_field_compare}
\end{figure}

\begin{figure}[t]
  \centering
  \includegraphics[page=1,width=\linewidth]{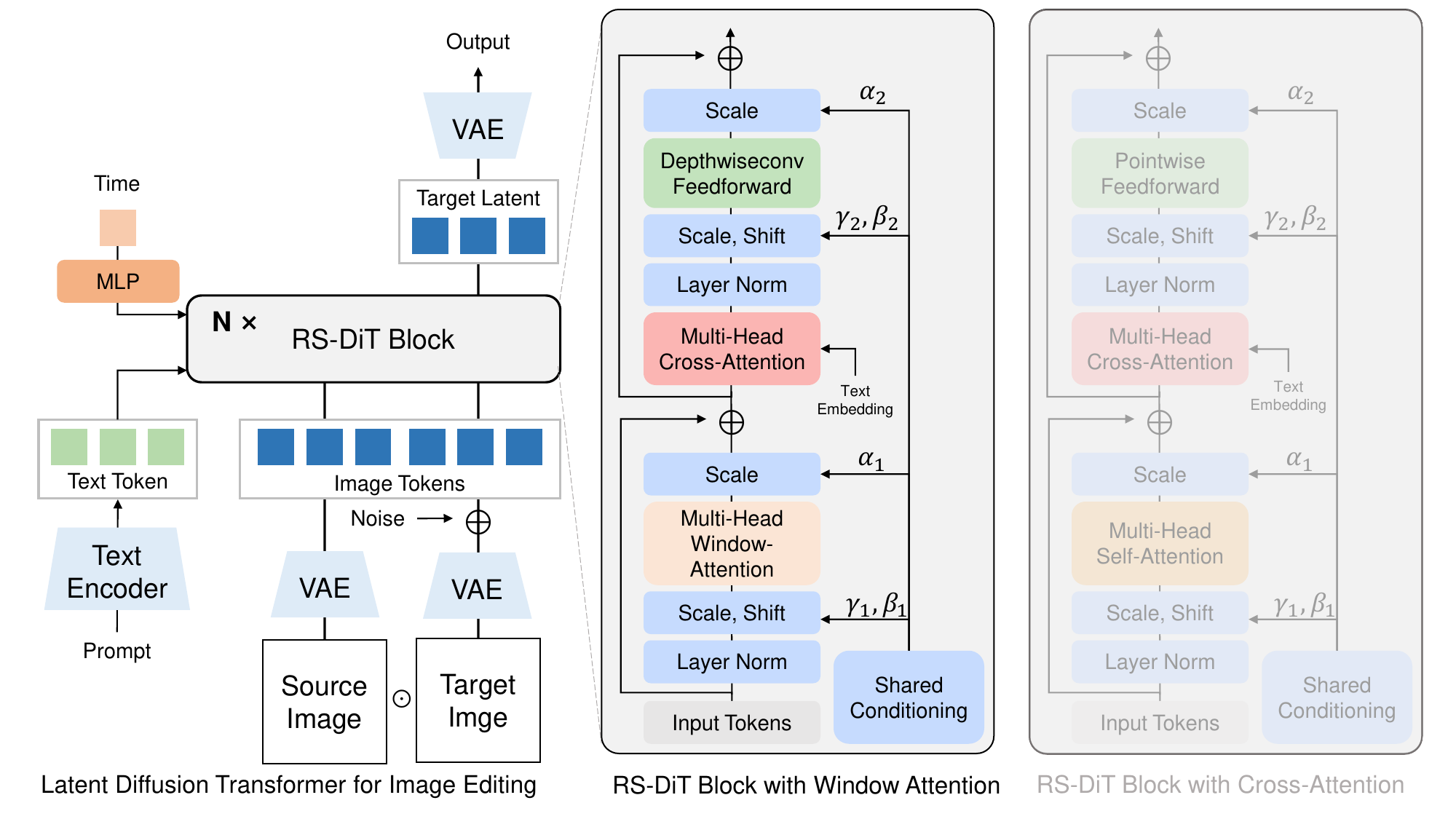}
  \caption{Architecture of the \rseditsubditplus{} variant.}
  \label{fig:rsedit_dit_plus}
\end{figure}

Figure~\ref{fig:receptive_field_compare} compares receptive fields at a query location (star) for (a)~convolution, (b)~global self-attention, and (c)~window attention. We introduce a \rseditsubditplus{} variant (Fig.~\ref{fig:rsedit_dit_plus}). We replace the original multi-head self-attention with multi-head window attention, motivated by the observation that \textbf{remote sensing editing relies more on local spatial cues than global semantics}. This design reduces GFLOPs with negligible parameter overhead ($<0.1\%$). We further replace the pointwise FFN with a depthwise-convolutional FFN and remove the position embedding (we maintain the position embedding in \rseditdit{}).

\subsection{Training Details}
We train \rsedit{} on the RSCC dataset~\cite{chenRSCCLargeScaleRemote2025a} using Prodigy as the optimizer with global batch size 32 on NVIDIA A100 GPUs. All variants are trained for 30{,}000 steps. The U-Net is initialized from SD~1.5, replacing the original text encoder with DGTRS-CLIP-ViT-L-14~\cite{chenDGTRSDDGTRSCLIPDualGranularity2025a}. DiT-based variants are initialized from PixArt-$\alpha$~\cite{chenPixArtaFastTraining2024}.

%% file: sections/results.tex
\section{Results}\label{sec:results}

\begin{figure*}[t]
  \centering
  \includegraphics[width=\linewidth]{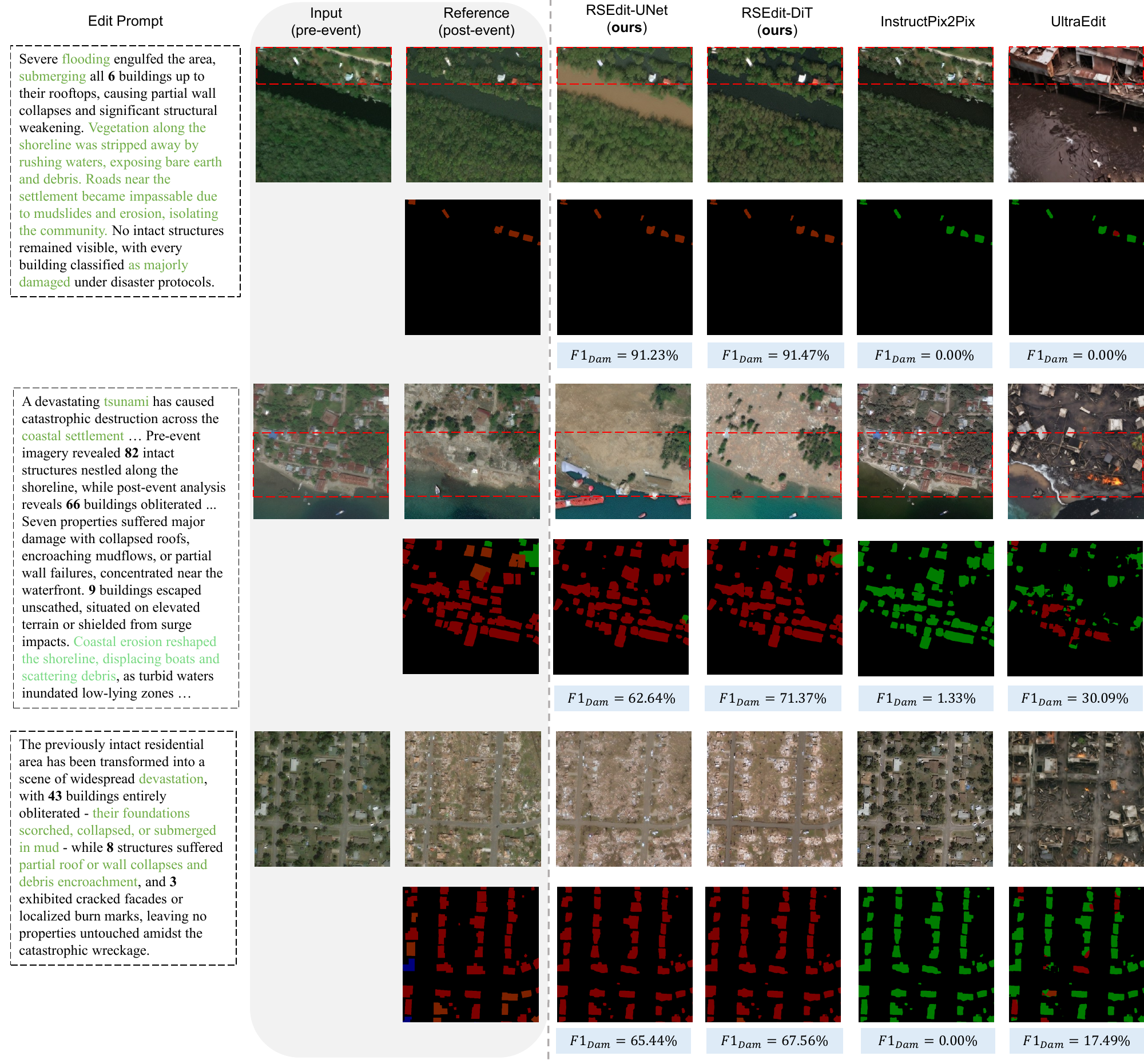}
  \caption{Qualitative comparison on disaster simulation. From left: Edit prompt, Input, Reference, \rseditsubunet, \rseditsubdit, InstructPix2Pix, UltraEdit. \rsedit produces physically plausible edits while baselines under-edit or introduce artifacts.}
  \label{fig:qualitative_results}
\end{figure*}

\begin{table}[h]
  \caption{Change detection evaluation results on RSCC. $F1_{\text{dam}}$ follows Eq.~\eqref{eq:f1dam} (xView2-style aggregation). \textbf{Bold} indicates best, \underline{underline} second best.}
  \label{tab:main_comparison}
  \centering
  \begingroup
  \setlength{\tabcolsep}{3pt}
  \small
  \begin{tabular*}{\linewidth}{@{\extracolsep{\fill}}lcccc}
    \toprule
    \multirow{2}{*}{Method} & F1\textsubscript{dam} $\uparrow$ & \multicolumn{3}{c}{VIE Scores $\uparrow$} \\
    \cmidrule(lr){3-5}
    & & SC & PQ & VIE \\
    \midrule
    SD~1.5$^\dagger$ & $6.57$ & $1.680$ & $5.104$ & $2.345$ \\
    SD~2.1$^\dagger$ & $7.90$ & $1.672$ & $4.479$ & $2.148$ \\
    Text2Earth$^\dagger$ & $15.28$ & $2.391$ & $5.003$ & $2.922$ \\
    \midrule
    InstructPix2Pix & $8.35$ & $3.435$ & $\underline{5.219}$ & $3.305$ \\
    MagicBrush & $0.94$ & $1.145$ & $4.642$ & $1.647$ \\
    UltraEdit & $1.13$ & $1.970$ & $4.935$ & $2.485$ \\
    Flux.1-Kontext & $8.44$ & $3.512$ & $4.962$ & $3.325$ \\
    \midrule
    \textbf{\rseditsubunet} & $\mathbf{49.74}$ & $\underline{4.050}$ & $\mathbf{5.358}$ & $\underline{4.123}$ \\
    \textbf{\rseditsubdit} & $30.28$ & $\mathbf{4.509}$ & $\underline{5.219}$ & $\mathbf{4.284}$ \\
    \textbf{\rseditsubditplus} & $\underline{45.35}$ & $3.604$ & $5.160$ & $3.698$ \\
    \bottomrule
  \end{tabular*}
  \endgroup
\end{table}

\subsection{Experimental Setup}
\noindent\textbf{Baselines.}
We compare against general instruction-based editors (InstructPix2Pix, MagicBrush, UltraEdit, and Flux-1-Kontext)~\cite{brooks2023instructpix2pix,zhangMagicBrushManuallyAnnotated2023,zhaoUltraEditInstructionbasedFineGrained2024b,labsFLUX1KontextFlow2025}. We also include RS-oriented baselines built from Text2Earth and SD with SDEdit~\cite{liuText2EarthUnlockingTextdriven2025,mengSDEditGuidedImage2022}.

\begin{figure}[t]
  \centering
  \includegraphics[width=\linewidth]{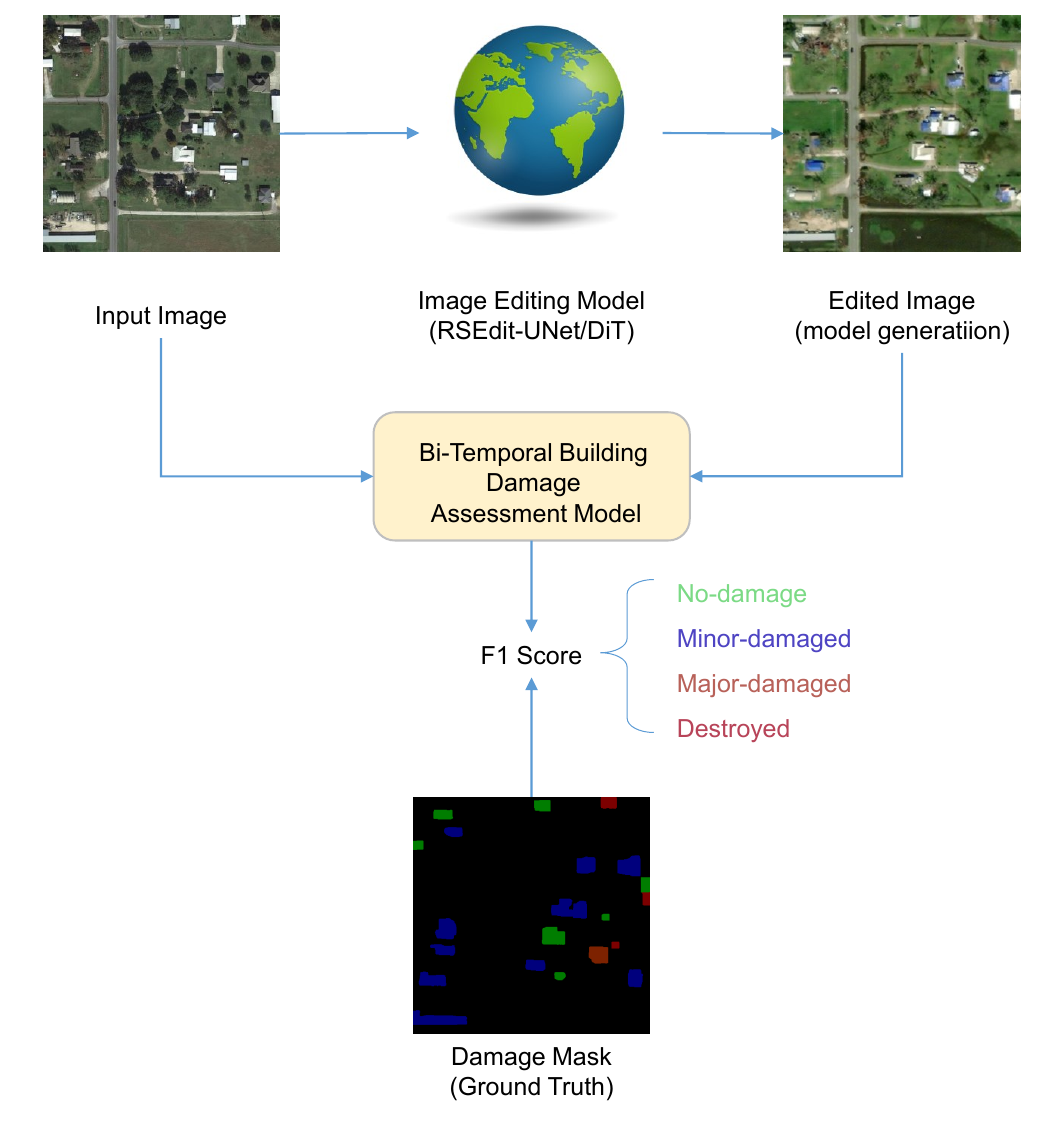}
  \caption{Proposed change-centric evaluation protocol. $F1_{\text{dam}}$ measures localized change correctness.}
  \label{fig:proposed_metrics}
\end{figure}

\noindent\textbf{Metrics.}
We report (i) $F1_{\text{dam}}$, our primary damage-aware change metric, using the evaluation protocol illustrated in Fig.~\ref{fig:proposed_metrics}. It is computed from a pre-trained building damage assessment model\footnote{Checkpoint from the EVER-Z Changen2-ChangeStar1x256 repository on Hugging Face.} on the xView2 dataset~\cite{guptaCreatingXBDDataset2019}.
For damage class $c \in \{1,2,3,4\}$ (none/minor/major/destroyed),
\begin{equation}
  F1_c=\frac{2\cdot TP_c}{2\cdot TP_c+FP_c+FN_c},
\end{equation}
aggregated as
\begin{equation}
  F1_{\text{dam}}=\frac{4}{\sum_{c=1}^{4}(F1_c+\epsilon)^{-1}}, \quad \epsilon=10^{-6}.
  \label{eq:f1dam}
\end{equation}
For sample-wise analysis we additionally use a pixel-weighted variant, $F1_{\text{weighted}}={\sum_{c=1}^{4} F1_c \cdot N_c}/{\sum_{c=1}^{4} N_c}$, where $N_c$ is the number of ground-truth pixels in class $c$.
We also report (ii) VIE~\cite{kuVIEScoreExplainableMetrics2024}: $VIE = \sqrt{SC \times PQ}$, where SC (Semantic Consistency) and PQ (Perceptual Quality) are VLM-rated on a 0--10 scale using \texttt{Qwen3.5-397B-A17B}.

\subsection{Main Results on RSCC}
Table~\ref{tab:main_comparison} summarizes quantitative comparisons under the above setup. On semantic change generation, \rsedit establishes a new state-of-the-art. \rseditsubunet achieves the highest F1\textsubscript{dam} of 49.74, with \rseditsubditplus as the second best at 45.35; both clearly outperform general-domain editors and RS-oriented baselines. \rseditsubdit achieves the highest SC (4.509) and the best overall VIE (4.284), indicating stronger instruction consistency. \rseditsubunet achieves the highest PQ (5.358), suggesting stronger perceptual quality. \rsedit variants consistently outperform baselines in overall VIE; the gap versus MagicBrush and UltraEdit highlights the necessity of domain-specific fine-tuning.

Qualitatively, \rsedit interprets complex geospatial instructions and produces physically plausible changes. In flooding scenarios, it correctly inundates low-lying regions while respecting elevation logic implied by building footprints. In tsunami simulations, it generates coherent coastal damage patterns. In building destruction cases, it removes roof textures and flattens structures as requested while preserving surrounding roads and vegetation. By contrast, InstructPix2Pix often under-edits, and UltraEdit tends to over-edit or introduce perspective distortions that violate orthographic constraints.

\subsection{Ablation Study on Conditioning Strategy}
We conduct a dedicated ablation over the conditioning layouts shown in Fig.~\ref{fig:conditioning_ablation}, comparing channel versus token fusion for both U-Net and DiT pipelines under the same editing setting.

\begin{table}[t]
  \caption{Ablation on conditioning strategy across U-Net and DiT backbones on RSCC. Columns are backbone/fusion pairs: U-Net with channel (ch.) or token (tk.) concatenation; DiT with token or channel fusion; DiT+ denotes \rseditsubditplus{} with token fusion.}
  \label{tab:conditioning_ablation}
  \centering
  \begingroup
  \setlength{\tabcolsep}{2.5pt}
  \footnotesize
  \begin{tabular*}{\columnwidth}{@{\extracolsep{\fill}}lccccc}
    \toprule
    & \multicolumn{2}{c}{U-Net} & \multicolumn{2}{c}{DiT} & DiT+ \\
    \cmidrule(lr){2-3}\cmidrule(lr){4-5}\cmidrule(lr){6-6}
    Metric & ch. & tk. & tk. & ch. & tk. \\
    \midrule
    \#Param & $\sim$860M & $\sim$860M & $\sim$610M & $\sim$610M & $\sim$610M \\
    GFLOPs & $\sim$1$\times$ & $\sim$2$\times$ & $\sim$3$\times$ & $\sim$1.5$\times$ & $\sim$2.5$\times$ \\
    F1\textsubscript{dam} $\uparrow$ & 49.74 & 38.29 & 30.28 & 41.37 & 45.35 \\
    SC $\uparrow$ & 4.050 & 4.024 & 4.509 & 4.337 & 3.604 \\
    PQ $\uparrow$ & 5.358 & 5.257 & 5.219 & 5.429 & 5.160 \\
    VIE $\uparrow$ & 4.123 & 3.988 & 4.284 & 4.267 & 3.698 \\
    \bottomrule
  \end{tabular*}
  \endgroup
\end{table}

Figure~\ref{fig:conditioning_ablation} and Table~\ref{tab:conditioning_ablation} summarize complementary views of the same design space; the latter quantifies clear strategy trade-offs. For DiT, channel concatenation improves $F1_{\text{dam}}$ from 30.28 to 41.37 and gives the highest PQ (5.429), while token concatenation retains the best SC (4.509) and VIE (4.284). \rseditsubditplus{} further pushes $F1_{\text{dam}}$ to 45.35 while trading off SC/VIE. For U-Net, channel concatenation matches the stronger \rseditsubunet{} configuration in Table~\ref{tab:main_comparison} on $F1_{\text{dam}}$ (49.74) and on SC/PQ/VIE, whereas token concatenation lowers all four metrics.

%% file: sections/conclusion.tex
\section{Conclusion}\label{sec:conclusion}

We presented \rsedit{} as a unified testbed for text-guided remote sensing image editing built from off-the-shelf latent diffusion, with explicit, backbone-specific conditioning of source and noisy latents. On RSCC we conducted a comprehensive study that jointly varies denoiser family (U-Net versus DiT), conditioning layout (channel versus token fusion), and a tailored \rseditsubditplus{} configuration, all under one training recipe and one evaluation protocol combining change-aware $F1_{\text{dam}}$, VIE-style instruction and perceptual scores, qualitative comparisons, and conditioning ablations. The paper records this design space, experimental setup, and released artifacts to support reproducible comparisons in RS editing.

\begin{figure*}[!b]
  \centering
  \includegraphics[width=\textwidth]{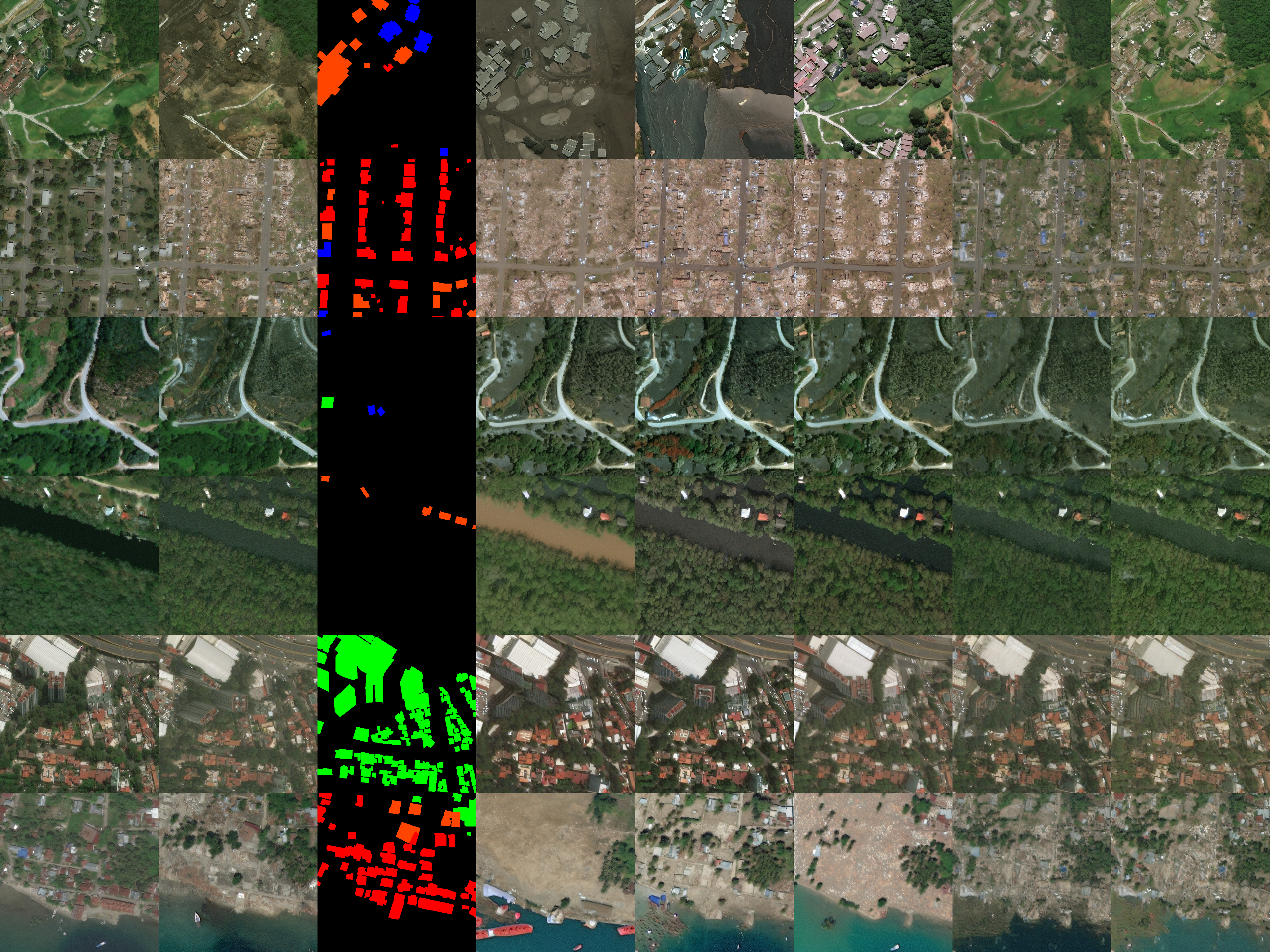}
  \caption{Additional qualitative comparison (six exemplar rows). Columns left to right: pre-event (GT), post-event (GT), GT color-coded change mask, \rseditsubunet{} with channel concatenation, \rseditsubunet{} with token concatenation, \rseditsubdit{} with token concatenation, \rseditsubdit{} with channel concatenation, and \rseditsubditplus{} (token concatenation).}
  \label{fig:qualitative_ablation_grid}
\end{figure*}